\documentclass{article}

\usepackage{PRIMEarxiv}

\usepackage[utf8]{inputenc} % allow utf-8 input
\usepackage[T1]{fontenc}    % use 8-bit T1 fonts
\usepackage{hyperref}       % hyperlinks
\usepackage{url}            % simple URL typesetting
\usepackage{booktabs}       % professional-quality tables
\usepackage{amsfonts}       % blackboard math symbols
\usepackage{amsmath}
\usepackage{nicefrac}       % compact symbols for 1/2, etc.
\usepackage{microtype}      % microtypography
\usepackage{lipsum}
\usepackage{fancyhdr}       % header
\usepackage{graphicx}       % graphics
\usepackage{subfigure}
\usepackage{caption} % for customizing table captions
\usepackage{siunitx}
\usepackage{float}
\usepackage{tabularx}
\graphicspath{{media/}}     % organize your images and other figures under media/ folder
\usepackage{algorithm}
\usepackage{algorithmic}
\usepackage{listings}
\usepackage{xcolor}

\hypersetup{
    colorlinks=true,
    linkcolor=blue,
    urlcolor=cyan,
}

\fancypagestyle{firstpage}{
    \fancyhf{}  % Clear all header/footer content
    \fancyfoot[L]{GitHub Repository: \href{https://github.com/SkAndMl/gpt-variations}{gpt-variations}}
}
\title{Towards Smaller, Faster Decoder-Only Transformers: Architectural Variants and Their Implications
}

\author{
  Sathya Krishnan Suresh \\
  Nanyang Technological University \\
  \texttt{sathyakr001@e.ntu.edu.sg} \\
   \And
  Shunmugapriya P \\
  Puducherry \\
  \texttt{pshunmugapriya@gmail.com}\\
}

\begin{document}
\maketitle

\thispagestyle{firstpage}

\begin{abstract}
In recent times, the research on Large Language Models (LLMs) has grown exponentially, predominantly focusing on models underpinned by the transformer architecture, and further developed through the decoder-only models such as Large Language Models (LLM). Contemporary efforts in this field primarily aim to enhance model capabilities by scaling up both the architecture and data volumes utilized during training. However, there has been little exploration into reducing model sizes while maintaining their effectiveness. In this study, we introduce three modifications to the decoder-only transformer architecture—namely ParallelGPT (\textit{pgpt}), LinearGPT (\textit{lgpt}), and ConvGPT (\textit{cgpt}). These variants demonstrate comparable performances to the conventional architecture, with \textit{lgpt} outperforming it in \textbf{4 out of 7 benchmarks} with \textbf{less than half the parameters}. We open-source the model weights and the complete codebase for these implementations for further research.
% \footnote{\url{https://github.com/SkAndMl/gpt-variations}}.
\end{abstract}

% keywords can be removed

\section{Introduction}

Since the debut of ChatGPT, there has been a notable increase in research on Large Language Models (LLMs) across a broad range of disciplines, made possible by the accessibility of this technology to a diverse user base. This fastly growing field has largely pursued two distinct paths: one aims at either scaling the model size or the training dataset (or both) to enhance performance, while the other concentrates on refining smaller models (ranging from 1B to 7B parameters) with high-quality data. Despite these advances, investigations into the structural modifications of the transformer architecture itself have been relatively overlooked. Recent studies challenge the necessity of perpetually increasing model sizes by demonstrating that the deeper layers of LLMs may have minimal influence on predictive outcomes \cite{gromov2024unreasonable}. 

In this work, we explore modifications to the decoder-only transformer architecture to address current challenges in the scalability and practical application of Large Language Models (LLMs). Recognizing the significant impact of model size on the computational overhead of training and inference, we introduce three variants—ParallelGPT (\textit{pgpt}), LinearGPT (\textit{lgpt}), and ConvGPT (\textit{cgpt})—each designed to reduce parameter count while maintaining, or potentially enhancing, model performance.

The models presented in this paper range from 30M to 80 parameters.  Our decision to train models of these sizes was influenced by GPT-2 \cite{gpt2}, which marked the beginning of the GPT era. We aimed to introduce architectural variations within the vicinity of GPT-2’s size to explore similar performance characteristics. Rather than striving for state-of-the-art results, our goal was to develop models with comparable parameters but reduced size, achieving performance similar to the traditional GPT architecture.

We pre-trained each architectural variant on the 10 billion token subset of the fineweb-edu \cite{fineweb_edu} dataset. The dataset was chosen to match the scale of the models we wanted to train.  By training on a reasonably extensive dataset, we aimed to ensure that the models could be effectively evaluated on various benchmarks, thereby comparing the performance of different architectures.
 
The remainder of this paper is organized as follows. The related works  are discussed in Section 2. The modifications made to the base architecture and the reasons for the modifications are presented in Section 3. Experimental setup and the results are presented in Section 4. Section 5 discusses the limitations of the research and finally Section 6 contains the conclusion of our work.

\begin{figure}[t]
  \centering
  \includegraphics[width=0.4\textwidth]{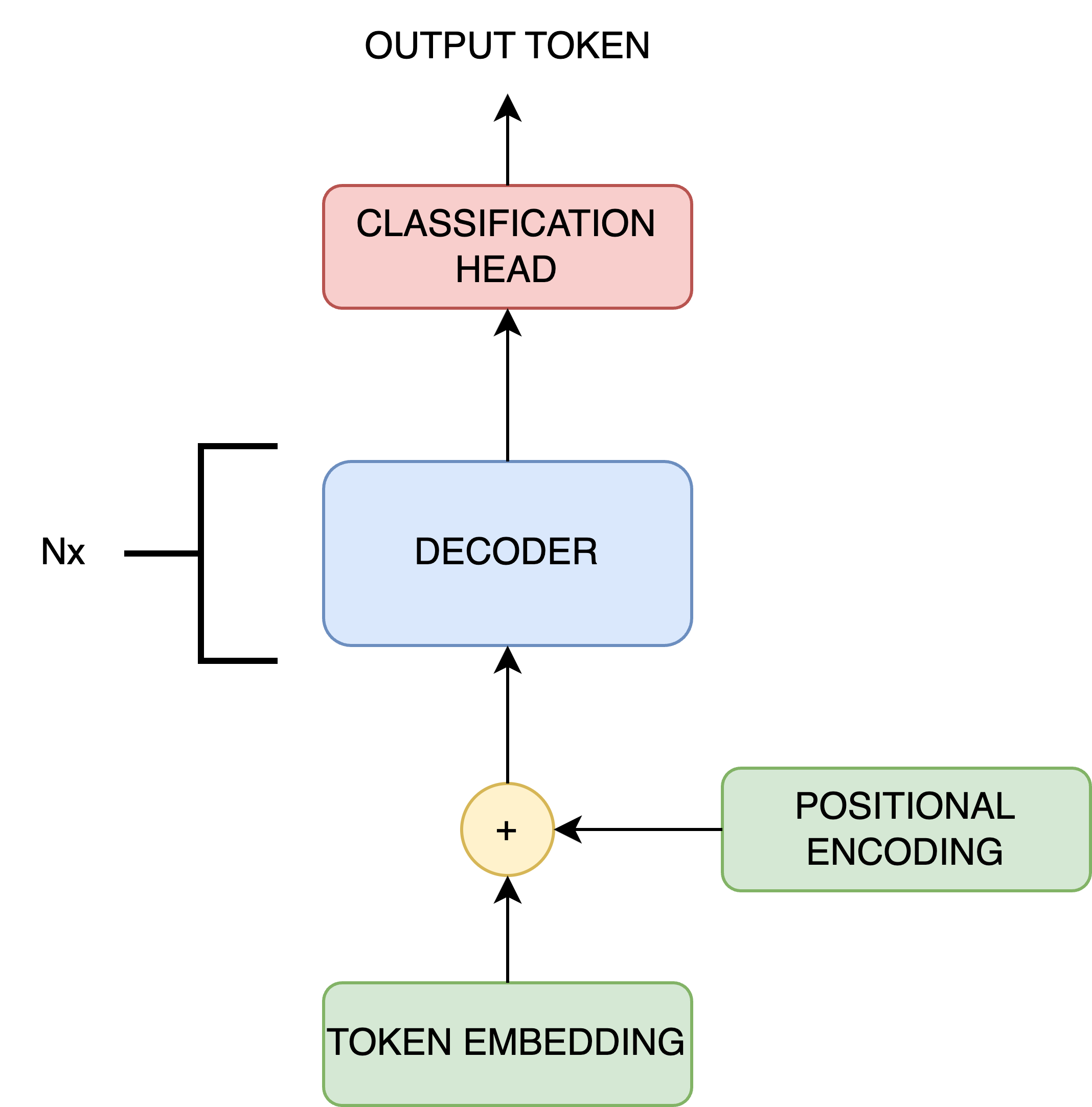}
  \caption{Traditional GPT architecture}
  \label{fig:gpt_arch}
\end{figure}

\section{Related Work}

The evolution of Large Language Models (LLMs) has been heavily influenced by advances in transformer architecture, first introduced by Vaswani et al. \cite{vaswani2017attention} and later adapted into decoder-only models such as GPT \cite{radford2018improving}. These models typically focus on scaling up, using more complex architectures and larger datasets to improve performance. However, there is growing interest in designing models that maintain strong performance while being smaller and more efficient, addressing the need for models that are simpler to train and deploy.

Our work introduces three new variants of the decoder-only transformer architecture aimed at improving training and inferencing efficiency: ParallelGPT (\textit{pgpt}), LinearGPT (\textit{lgpt}), and ConvGPT (\textit{cgpt}). These models achieve similar performance to traditional transformer architectures in text generation tasks but with fewer parameters and faster training times. We have open-sourced the model weights and codebase to support further research.

Recent research also explores efficiency-focused architectures. For example, the Funnel-Transformer \cite{dai2020funnel} reduces computational costs by compressing the sequence of hidden states through a pooling mechanism, allowing the model to expand in depth or width without increasing the overall cost.

The MobiLlama model \cite{thawakar2024mobillama} is another example, optimized for use in resource-constrained settings. It uses a parameter-sharing strategy within transformer blocks to minimize both pre-training and deployment costs, aligning with the goal of creating smaller, more efficient models.

Innovative attention mechanisms have also contributed to more efficient LLMs. Grouped Query Attention \cite{ainslie2023gqa} reduces the complexity of attention calculations by grouping inputs, effectively lowering the computation from quadratic to linear, which helps in handling longer sequences. Multi-Query Attention \cite{shazeer2019transformer} further improves efficiency by allowing multiple queries within a single attention head, enhancing the model’s ability to gather and process diverse information.

These advances reflect a clear trend toward making LLMs not just more powerful but also more adaptable to practical constraints like size and computational limits. Our work builds on this trend by exploring novel architectural modifications that offer a balance between performance and efficiency.

\section{Architectural modifications}
In this section, we introduce three novel architectures derived from the traditional GPT architecture to address various limitations in training and inference. These architectures are designed to enable faster training and inference. The three proposed architectures are ParallelGPT (\textit{pgpt}), LinearGPT (\textit{lgpt}), and ConvGPT (\textit{cgpt}).

\subsection{ParallelGPT}
In a traditional GPT architecture, the $N$ decoder blocks of the same dimensionality are stacked on top of each other. The drawback of such an architecture is that as \textit{N} increases, the deeper blocks have little information to work with. Also recent research \cite{gromov2024unreasonable} point out the inefficiency of the deeper layers thereby questioning the need for increasing $N$. To overcome these drawbacks, we propose splitting the $N$ decoder blocks into \textit{P} parallel blocks such that each parallel path $P_i$ consists of $N/P$ decoder blocks. Each \textit{$P_i$} has a preceding dense layer $PD_i$ to process the incoming token embeddings. The reason for introducing a dense layer before each parallel block is to make sure that each block learns something different. A vector of learnable weights $W \in \mathbb{R}^P$ is used to combine the outputs of each of the parallel blocks. The following set of equations give the mathematical representation of the architecture.

\begin{equation}\label{eq:pgpt_1}
    h_i = PD_i(x), \quad \text{for } i = 1, 2, \ldots, P
\end{equation}

\begin{equation}\label{eq:pgpt_2}
    y_i = f_i(h_i), \quad \text{for } i = 1, 2, \ldots, P
\end{equation}

\begin{equation}\label{eq:pgpt_3}
    \alpha_i = \frac{\exp(w_i)}{\sum_{j=1}^{P} \exp(w_j)}, \quad \text{for } i = 1, 2, \ldots, P
\end{equation}
\begin{equation}\label{eq:pgpt_4}
    y = \sum_{i=1}^{P} \alpha_i \cdot y_i
\end{equation}

We designed this architecture so that we can have the following benefits:
\begin{itemize}
    \item Faster training, as each parallel block $P_i$ can be assigned to a separate compute node, allowing them to be trained in parallel.
    \item During inference, we have the flexibility to drop 0 to $P - 1$ number of blocks based on the required inference speed, trading off some output quality for faster performance.
\end{itemize}
A point to note here is that we have used a simple weighting mechanism for combining the outputs but other methods like mixture of experts, gating functions can also be used, which we leave for future research. 

\begin{figure*}[t]
  \centering
  \includegraphics[width=0.6\textwidth]{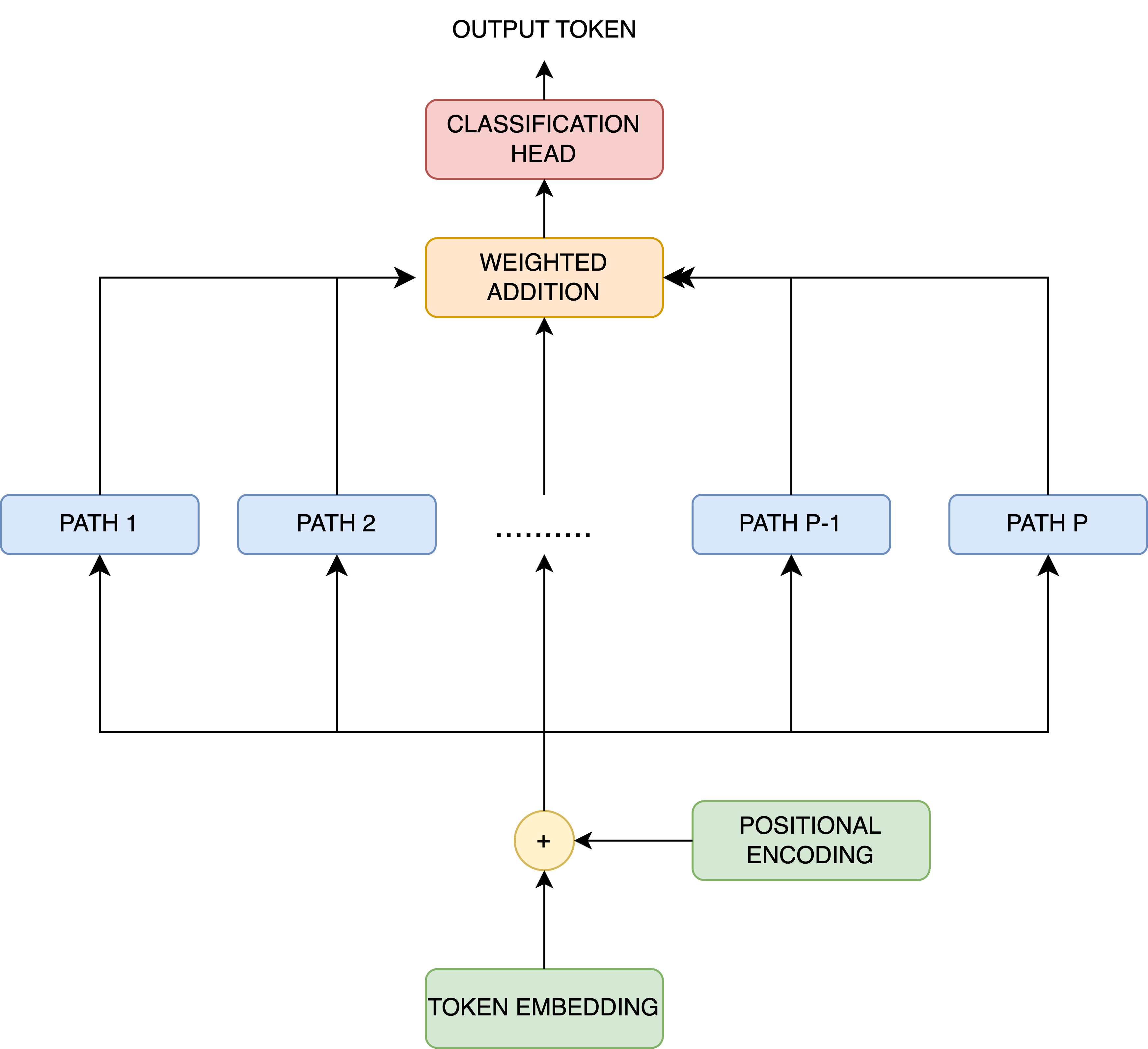}
  \hfill
  \caption{ParallelGPT}
\end{figure*}

\begin{figure*}[t]
  \centering
  \includegraphics[width=0.4\textwidth]{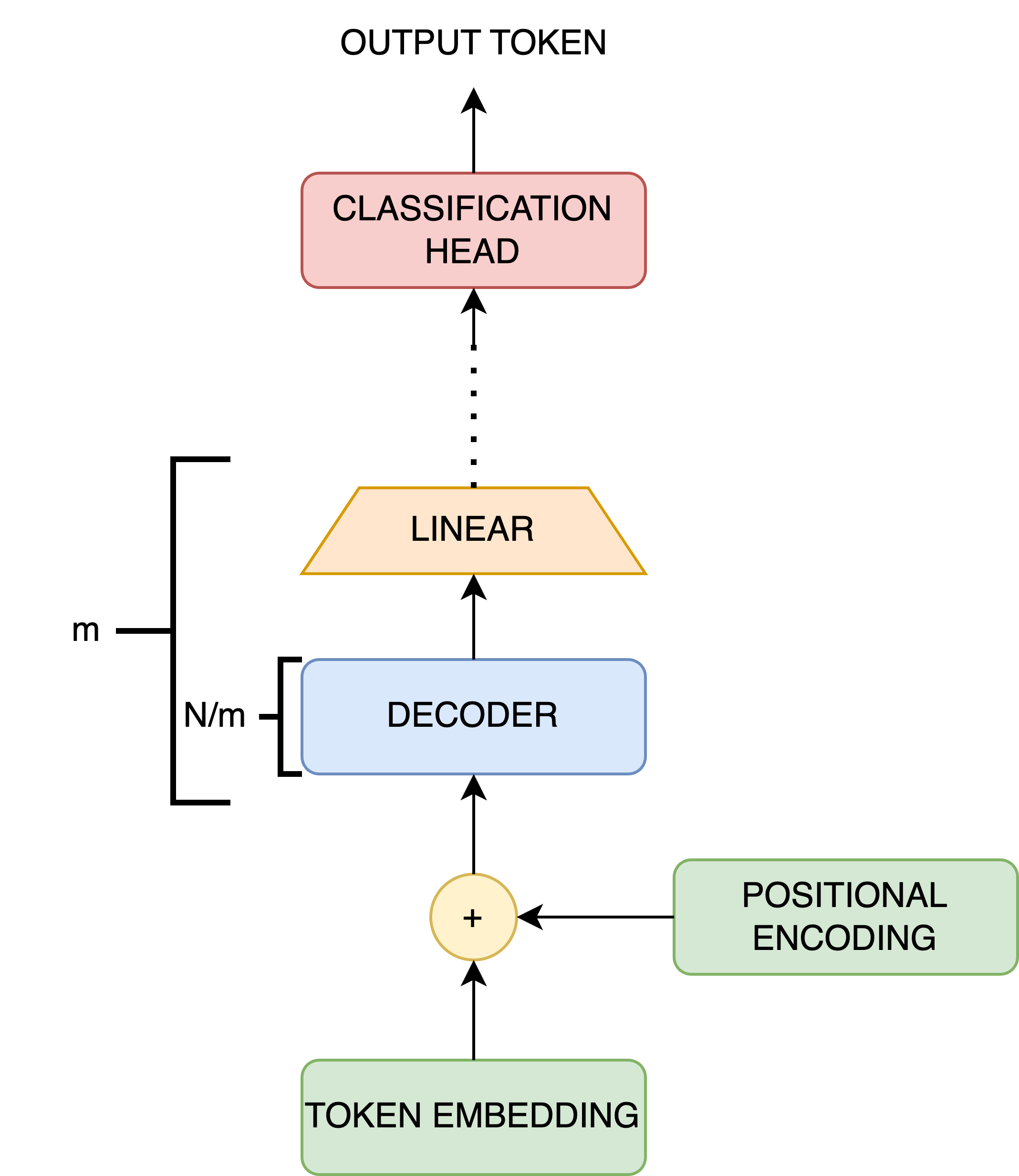}
  \hfill
  \includegraphics[width=0.4\textwidth]{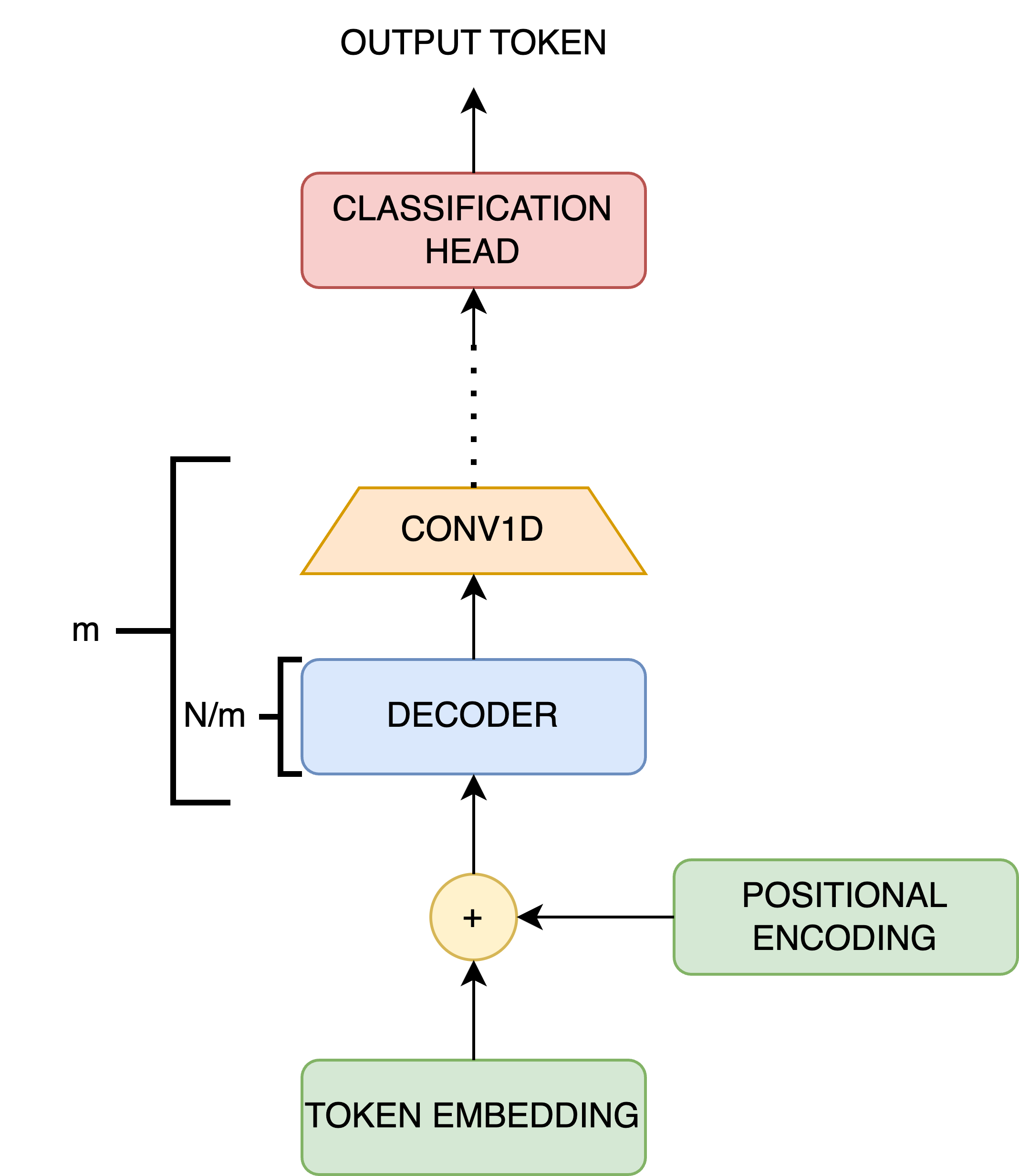}
  \caption{LinearGPT and ConvGPT}
\end{figure*}

\subsection{LinearGPT}
As discussed in \cite{gromov2024unreasonable}, the earlier layers contribute more to the generation capabilities of an LLM than the deeper layers. This questions the need for each decoder block having the same dimension as this results in the model having much more parameters than that is actually needed for the desired performance. The extra parameters of the model might also result in the model overfitting to the training set which LLMs are known to do. 

To address these issues, we propose an architecture, \textit{lgpt}, in which the dimension of the decoder blocks keeps reducing by half as the embedding of a token passes through the architecture. This concept is inspired from the continuous downsampling of images in CNNs for a task like image classification. To make sure that the dimension of the embedding of a token does not get too small by the time it reaches the classification head, we only reduce the dimension after every $n$ number of decoder blocks. This is done by introducing a dense or a linear layer after every $n$ decoder blocks. The following equations \ref{eq:lgpt_eq1}, \ref{eq:lgpt_eq2} and \ref{eq:lgpt_eq3} give a mathematical representation for the architecture discussed. Here $D_i$ represents the $i^{th}$ decoder block in an $N$ number of stacked decoder blocks and Eq. \ref{eq:lgpt_eq3} gives us the dimension $d_m$ of a vector before it goes through a decoder block.

\begin{equation}\label{eq:lgpt_eq1}
\mathbf{h}_i = \text{D}_i(\mathbf{h}_{i-1}), \quad i = 1, 2, \ldots, N
\end{equation}

\begin{equation}\label{eq:lgpt_eq2}
\mathbf{h}_{i} = 
\begin{cases}
\text{L}_j(\mathbf{h}_{i-1}), & \text{if } i = nj \\
\mathbf{h}_{i-1}, & \text{otherwise}
\end{cases}
\end{equation}

\begin{equation}\label{eq:lgpt_eq3}
d_m = \frac{d_0}{2^m}, m = \left\lfloor \frac{N}{n} \right\rfloor
\end{equation}

The \textit{lgpt} architecture can reduce the number of parameters in the model in terms of millions or billions. The reduction potential of this architecture is shown in the Appendix \ref{app:reduction_potential}. In the experiments we did, the number of parameters in the traditional \textit{gpt} architecture was \textbf{77.2M} parameters while the \textit{lgpt} architecture had only \textbf{36.4M} parameters and it still performed competitively with \textit{gpt}, while outperforming it COPA and ARC Easy \cite{clark2018arc} as shown in Table \ref{tab:benchmark_comparison}.

\subsection{ConvGPT}
A one-dimensional convolution with kernel size and stride as 1 is a very similar operation when compared with the linear operation as both of them have the same number of parameters to train and both of them carry out a matrix multiplication of similar dimensions but at a fundamental level both are different because of the positional context, spatial invariance and weight sharing properties possessed by the convolutional operations. In order to find whether using one dimensional convolution for compressing the dimensions, has a performance difference compared to the traditional \textit{gpt} and the previous architectures introduced, we introduce a third new architecture \textit{cgpt} which replaces the linear compression layers in \textit{lgpt} with a one dimensional convolution layer. \textit{cgpt} offers the same reduction in number of parameters as \textit{pgpt} and as can be seen from Table \ref{tab:benchmark_comparison} it is competitive with the traditional \textit{gpt} architecture.

\section{Experimental Setup and Results}
This section discusses the specific parameters we used for the experiments, the training parameters used and the results of each architecture on various benchmarks.

\subsection{Model Configuration}
For each of the architecture the common hyperparameters are set as illustrated in Table \ref{tab:gpt_config}. For the \textit{pgpt} architecture, we set the number of parallel blocks $P$ in this experiment to 2 with each parallel path $P_i$ having 4 decoder blocks. For the \textit{lgpt} and \textit{cgpt} architectures we set $n$ (number of blocks after which a linear or convolutional layer is introduced) as 2 resulting in the final embedding dimension to be 32. The number of parameters and the memory requirements for each of the models are shown in Table \ref{tab:model_hardware}, where \textit{pgpt-1}, refers to a \textit{pgpt} model in which the one parallel block is dropped. Note that during training, none of the parallel blocks are dropped. During inferencing and saving the model, 0 to $P - 1$ blocks can be dropped to save time and memory. We have algorithmically illustrated the reduction potential of the \textit{pgpt} architecture in Appendix \ref{app:reduction_potential_pgpt}

\begin{table}[h]
\centering
\begin{tabular}{|l|l|}
\hline
\textbf{Parameter}       & \textbf{Value} \\ \hline
\texttt{context\_size}   & 1024           \\ \hline
\texttt{vocab\_size}     & 50304          \\ \hline
\texttt{n\_layer}        & 8              \\ \hline
\texttt{n\_head}         & 8              \\ \hline
\texttt{embedding\_dimension} & 512        \\ \hline
\texttt{dropout}         & 0.0            \\ \hline
\end{tabular}
\caption{Common Configuration Parameters}
\label{tab:gpt_config}
\end{table}

\begin{table}[h]
\centering
\begin{tabular}{|l|l|l|}
\hline
\textbf{Model}       & \textbf{\# Params} & \textbf{Memory} \\ \hline
\texttt{gpt}         & 77.2M             & 294.53MB            \\ \hline
\texttt{pgpt}        & 77.7M             & 296.53MB            \\ \hline
\texttt{lgpt}        & 36.4M             & 138.95MB            \\ \hline
\texttt{cgpt}        & 36.4M             & 138.95MB            \\ \hline
\texttt{pgpt-1}      & 64.8M             & 247.52MB            \\ \hline
\end{tabular}
\caption{Model Hardware Requirements}
\label{tab:model_hardware}
\end{table}

\subsection{Training}
We train all the models in the same training setup which we have listed in Table \ref{tab:training_parameters}. We use a cosine decay for the learning rate. We train each model for 5000 steps which is about 25\% percent of the entire training set and the training was stopped here as due to compute constraints. We train each model on \texttt{4xA5000} GPUs, where the \textit{gpt} and \textit{pgpt} architectures took about 2 hours to train while \textit{lgpt} and \textit{cgpt} architectures took about 1 hour to train. 

\begin{table}[h]
    \centering
    \begin{tabular}{|c|c|}
        \hline
        \textbf{Hyperparameter} & \textbf{Value} \\ \hline
        \texttt{Batch size}     &  16            \\ \hline
        \texttt{Gradient accum. steps} & 8       \\ \hline
        \texttt{Min learning rate} & 6e-5        \\ \hline
        \texttt{Max learning rate} & 6e-4        \\ \hline
        \texttt{Max steps}        &  19000       \\ \hline
        \texttt{Warmup steps}     &  700         \\ \hline
        \texttt{Weight decay}     &  1e-1        \\ \hline
    \end{tabular}
    \caption{Training parameters}
    \label{tab:training_parameters}
\end{table}

\subsection{Results}

In this subsection, we present the loss curves and the evaluation results of our proposed model variants, comparing their performance on the following benchmarks

\begin{itemize}
    \item \textbf{HellaSwag: \cite{zellers2019hellaswag}} Tests the model's ability to understand commonsense reasoning and continuation of textual descriptions, evaluating how well the model predicts the next plausible sentence in narrative-like sequences.
    
    \item \textbf{WinoGrande: \cite{sakaguchi2019winogrande}} Measures the model's performance on pronoun resolution tasks, which challenges the model's understanding of context and disambiguation of pronouns in complex sentences.
    
    \item \textbf{CommonSenseQA: \cite{talmor2019commonsenseqa}} Focuses on evaluating the model's general commonsense knowledge and reasoning abilities, particularly in answering multiple-choice questions based on everyday scenarios.
    
    \item \textbf{ANLI (Adversarial Natural Language Inference): \cite{nie2019adversarial}} Tests the model's robustness and understanding of natural language inference under adversarial conditions, requiring the model to distinguish between entailment, contradiction, and neutral statements.
    
    \item \textbf{PIQA (Physical Interaction Question Answering): \cite{bisk2020piqa}} Evaluates the model's grasp of physical commonsense reasoning, specifically how well it can predict plausible actions or outcomes related to everyday physical interactions.
    
    \item \textbf{COPA (Choice of Plausible Alternatives): \cite{gordon-etal-2012-semeval}} Assesses the model’s ability to infer causality and select the more plausible of two alternative outcomes given a premise, testing causal reasoning skills.
    
    \item \textbf{ARC Easy: \cite{clark2018arc}} Tests scientific knowledge and reasoning in a simplified setting, focusing on how well the model can answer multiple-choice questions related to elementary and middle school science topics.
\end{itemize}

\begin{table*}[h]
\centering
\resizebox{0.7\textwidth}{!}{%
\begin{tabular}{|l|c|c|c|c|c|}
\hline
\textbf{Benchmark}    & \textbf{\textit{gpt}} & \textbf{\textit{pgpt}} & \textbf{\textit{lgpt}} & \textbf{\textit{cgpt}}  & \textbf{\textit{pgpt-1}} \\ \hline
HellaSwag             & \textbf{0.2625}       & 0.2604        & 0.2517        & 0.2492  & 0.2527      \\ \hline
WinoGrande            & \textbf{0.5036 }      & 0.4925        & 0.4870        & 0.4751    & 0.4933    \\ \hline
CommonSenseQA         & 0.1376       & \textbf{0.1605}        & 0.1458        & 0.1540    & 0.1572    \\ \hline
ANLI                  & 0.3300       & \textbf{0.3350}        & 0.3290        & 0.3240   & 0.3340     \\ \hline
PIQA                  & 0.5071       & 0.5131        & 0.5185        & \textbf{0.5256 } & 0.5125      \\ \hline
COPA                  & 0.5300       & 0.5200        & 0.5500        & 0.5000   & \textbf{0.5600}     \\ \hline
ARC Easy              & 0.2333       & 0.2439        & \textbf{0.2491}        & 0.2298   &  0.2509     \\ \hline
\end{tabular}%
}
\caption{Benchmark Results for Different Models}
\label{tab:benchmark_comparison}
\end{table*}

Table \ref{tab:benchmark_comparison} shows the performance of the various proposed architectures on a set of standard benchmarks. Note that the aim of this work is not to achieve state-of-the-art language models since the scale of the models and data used in this work is very small compared to the state-of-the-art models. The goal of this research is to produce architectural variants that are smaller in size with similar model parameters and have a performance comparable to the traditional \textit{gpt} architecture in the same training and evaluation environments.

\paragraph{Comparison with the traditional architecture:} The results from the table indicate that the \textit{lgpt}, \textit{cgpt} and \textit{pgpt-1} architectures perform competitively with the traditional \textit{gpt} architecture. The \textit{pgpt-1} outperforms \textit{gpt} on \textbf{5 out of the 7 benchmarks} while \textit{lgpt} and \textit{cgpt} outperform \textit{gpt} on \textbf{4 out of the 7 benchmarks} and \textbf{2 out of the 7 benchmarks} respectively. Even on benchmarks that \textit{gpt} performs the best, the other models perform competitively with \textit{gpt}. We hypothesize that the possible reason for the competitive performance of the architectural variants is that the reduced number of parameters acts as a regularization and reduces the likelihood of the model overfitting to the training data which is usually the case in the traditional decoder-only models which can be prompted to output its training data. The results further indicate that even with reduced number of parameters the architectural variants are competitive thereby asking the following questions which we believe future research will answer,
\begin{itemize}
    \item Which components of the decoder are most critical for generative capabilities?
    \item Is it necessary to scale uniformly across all decoder blocks, or can varied dimensional scaling offer similar or improved results?
    \item Should we test different architecture designs before scaling the models?
\end{itemize}

\paragraph{Comparison b/w \textit{pgpt} and \textit{pgpt-1}:} From the results it can be seen that \textit{pgpt-1} with just four decoder blocks is able to perform as well as the \textit{pgpt} model with eight decoder blocks. To further investigate if the network had learned to ignore the signal for one of the parallel paths and just used the signal from the other parallel path, we checked the final trained weight vector $W \in \mathbb{R}^2$. The weights of the parallel paths $P_1$ and $P_2$ were \texttt{0.5964} and \texttt{0.4036} respectively, which shows that the model used the signal from both the paths during training and the performance of the architecture remained robust during inference even when $P_2$ was dropped. The results raise the following questions for future research,
\begin{itemize}
    \item Does each parallel path model different part of the knowledge and if so can an efficient mechanism be designed process different sequences only with the required parallel paths?
    \item Can the \textit{pgpt} architecture be leveraged to introduce domain knowledge more effectively compared to post-pretraining methods like LoRA \cite{hu2022lora}? Specifically, can pgpt's unique parallel paths enable the direct integration of domain-specific knowledge during the pretraining stage by assigning dedicated blocks to learn varied domain aspects, or through the addition of specialized blocks post-pretraining, thus simplifying the process and enhancing model adaptability?
\end{itemize}

\begin{figure*}[t]
  \includegraphics[width=\linewidth]{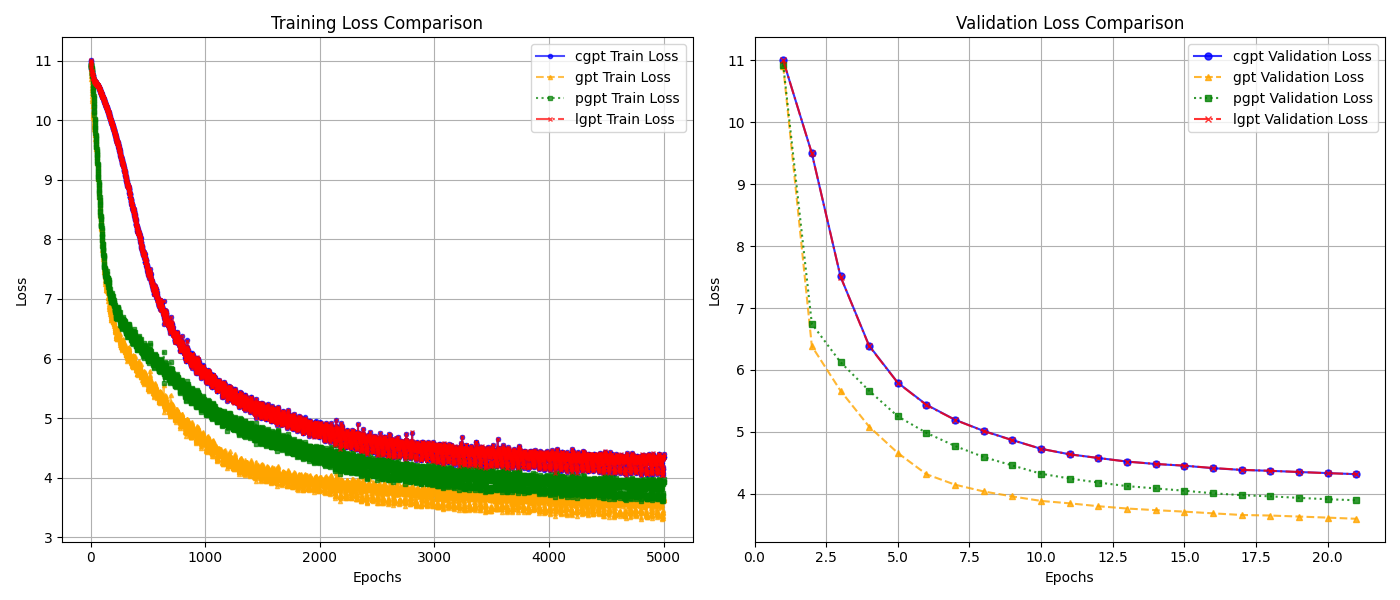} \hfill
  \caption {Loss comparison b/w the 4 models}
\end{figure*}

\section{Conclusion}

In this work, we presented three architectural variants of the traditional GPT architecture, aiming to investigate whether models with fewer parameters can achieve performance levels comparable to the original GPT. Our results across various benchmarks indicate that this is a promising and underexplored area of research. Through our findings, we raise critical questions and emphasize the need for further exploration. We believe that advancing this line of research is crucial, as AI can only reach its full potential to benefit humanity when it becomes widely accessible—a goal hindered by the high training and inference costs of large models today.

\section{Limitations}

While our proposed architectures demonstrate promising results, several limitations should be noted:

\begin{itemize}
    \item \textbf{Lack of Comparative Analysis with SOTA Models:} The architectural variants have not been directly compared to state-of-the-art (SOTA) models due to resource constraints. While our research focuses on architectures of GPT-2 scale, direct comparisons with SOTA models could provide critical insights into the performance and efficiency of the proposed variants. 

    \item \textbf{Computational Resource Constraints:} Due to computational and scale limitations, the training of our models was stopped after what we determined to be a reasonable number of steps. Further experiments with extended training durations on larger compute setups are necessary to explore the limits of these architectures.

    \item \textbf{Scalability and Applicability:} The scalability of the proposed architectures to larger model sizes and datasets remains untested. While effective at smaller scales, the performance and efficiency of \textit{pgpt}, \textit{lgpt}, and \textit{cgpt} in large-scale, real-world applications need further investigation. Future work should explore how these models adapt when scaled up or deployed in complex scenarios.

    \item \textbf{Impact of Simplified Weighting Mechanisms:} The use of simple weighting for combining outputs from parallel paths in \textit{pgpt} may not fully exploit the architecture's potential. Future research could explore more sophisticated methods, such as gating mechanisms or mixture of experts, to enhance the model's performance and adaptability.
\end{itemize}

Despite these limitations, our work provides valuable insights into architectural innovations in transformer models and opens new avenues for future research. 

\nocite{*}
%Bibliography
\bibliographystyle{unsrt}  
\bibliography{references}  
\clearpage
\appendix
\section{Reduction Potential of the \textit{lgpt} Architecture}
\label{app:reduction_potential}

In this section, we illustrate the reduction potential possessed by the \textit{lgpt} architecture. The python code given below illustrates the algorithm we used for calculating the number of parameters in the traditional \textit{gpt} architecture and the \textit{lgpt} architecture. 

\lstset{
  language=Python,
  basicstyle=\ttfamily\footnotesize,
  keywordstyle=\color{blue}\bfseries,
  commentstyle=\color{green!50!black},
  stringstyle=\color{red!70!black},
  showstringspaces=false,
  numbers=left,
  numberstyle=\tiny\color{gray},
  stepnumber=1,
  numbersep=5pt,
  backgroundcolor=\color{gray!10},
  frame=single,
  breaklines=true,
  breakatwhitespace=true,
  tabsize=4,
  captionpos=b
}

\begin{lstlisting}[caption={Calculation of Parameters and Reduction for GPT and LGPT Models}]
def calculate_params_and_reduction(config, layer_red=2):
    # GPT Parameters Calculation
    embedding_params = config.vocab_size * config.n_embd
    block_params = config.n_layer * (12 * config.n_embd**2 + (config.n_embd if config.bias else 0))
    classification_head_params = config.n_embd * config.vocab_size
    gpt_params = embedding_params + block_params + classification_head_params

    # LGPT Parameters Calculation
    total_lgpt_params = embedding_params
    current_embd = config.n_embd

    for i in range(config.n_layer):
        linear_layer_params = 0
        if i % layer_red == 0 and i != 0:
            current_embd = max(1, current_embd // 2)
            linear_layer_params = current_embd * 2 * current_embd
        block_params = 12 * current_embd**2 + (current_embd if config.bias else 0)
        total_lgpt_params += block_params + linear_layer_params

    classification_head_params = current_embd * config.vocab_size
    lgpt_params = total_lgpt_params + classification_head_params

    # Reduction Calculation
    reduction_frac = (gpt_params - lgpt_params) / gpt_params

    return gpt_params, lgpt_params, reduction_frac
\end{lstlisting}

We count the parameters in both the \textit{gpt} and the \textit{lgpt} model the parameters as following: the vocab size is 50304, context length is 1024, embedding dimension is 1600, number of layers is 48 and the number of heads is 16. In the \textit{lgpt} architecture we reduce the dimension after every 12 blocks so that the embedding dimension does not become too small. The following results demonstrate the effectiveness of \textit{lgpt} in reducing the overall model size:

\begin{itemize}
    \item \textbf{gpt Parameters:} 1,635,532,800
    \item \textbf{lgpt Parameters:} 581,827,200
    \item \textbf{Reduction in Size:} 64.43\%
\end{itemize}

The results show that \textit{lgpt} architecture can reduce the number of parameters in the model by \textbf{more than a billion parameters} for the same configuration parameters. We leave the comparison of models of this scale for future research due to compute constraints.

\newpage

\section{Reduction Potential of the \textit{pgpt} Architecture}
\label{app:reduction_potential_pgpt}

In this section, we illustrate the reduction potential possessed by the \textit{pgpt} architecture. The python code given below illustrates the algorithm we used for calculating the number of parameters in the traditional \textit{gpt} architecture and the \textit{pgpt} architecture.

\begin{lstlisting}[caption={Calculation of Parameters and Reduction for GPT and PGPT Models}]

def calculate_params_and_reduction_pgpt(config, m_parallel_paths=8, m_drop=4):
    # GPT Parameters Calculation
    embedding_params = config.vocab_size * config.n_embd
    block_params = config.n_layer * (12 * config.n_embd**2 + (config.n_embd if config.bias else 0))
    classification_head_params = config.n_embd * config.vocab_size
    gpt_params = embedding_params + block_params + classification_head_params

    # PGPT Parameters Calculation
    total_pgpt_params = embedding_params
    params_per_layer = 12*config.n_embd*2 + (config.n_embd if config.bias else 0)
    block_params = config.n_layer * params_per_layer
    drop_block_params = m_drop*(config.n_layer//m_parallel_paths) * params_per_layer
    block_params -= drop_block_params
    projection_params = (m_parallel_paths-m_drop)*(config.n_embd*config.n_embd)
    classification_head_params = config.n_embd*config.vocab_size
    total_pgpt_params +=  block_params + projection_params + classification_head_params 

    # Reduction Calculation
    reduction_frac = (gpt_params - total_pgpt_params) / gpt_params

    return gpt_params, total_pgpt_params, reduction_frac

\end{lstlisting}

We count the parameters in both the \textit{gpt} and the \textit{pgpt} model the parameters as following: the vocab size is 50304, context length is 1024, embedding dimension is 1600, number of layers is 48 and the number of heads is 16. In the \textit{pgpt} architecture we introduce 8 parallel paths and calculate the parameters during inference assuming four parallel paths are dropped.The following results demonstrate the effectiveness of \textit{pgpt} in reducing the overall model size:

\begin{itemize}
    \item \textbf{gpt Parameters:} 1,635,532,800
    \item \textbf{pgpt Parameters:} 172,134,404
    \item \textbf{Reduction in Size:} 89.48\%
\end{itemize}

The results show that \textit{pgpt} architecture can reduce the number of parameters in the model by \textbf{almost a billion and a half parameters} for the same configuration parameters. We leave the comparison of models of this scale for future research due to compute constraints.

\end{document}